# TractGraphFormer: Anatomically Informed Hybrid Graph CNN-Transformer Network for Interpretable Sex and Age Prediction from Diffusion MRI Tractography

Yuqian Chen[a], Fan Zhang[b], Meng Wang[c], Leo R. Zekelman[d], Suheyla Cetin-Karayumak[e], Tengfei Xue[f], Chaoyi Zhang[f], Yang Song[g], Jarrett Rushmore[h], Nikos Makris[i], Yogesh Rathi[e], Weidong Cai[f], Lauren J. O'Donnell[a]

[a] Department of Radiology, Brigham and Women's Hospital, Harvard Medical School, Boston, MA, USA

[b] School of Information and Communication Engineering, University of Electronic Science and Technology of China, Chengdu, China

[c] Beth Israel Deaconess Medical Center, Harvard Medical School, Boston, MA, USA

[d] Department of Neurosurgery, Brigham and Women's Hospital, Harvard Medical School, Boston, MA, USA

[e] Department of Psychiatry, Brigham and Women's Hospital, Harvard Medical School, Boston, MA, USA

[f] School of Computer Science, The University of Sydney, Sydney, NSW, Australia

[g] School of Computer Science and Engineering, University of New South Wales, Sydney, NSW, Australia

[h] Departments of Anatomy and Neurobiology, Boston University School of Medicine, Boston, USA

[i] Departments of Psychiatry and Neurology, Massachusetts General Hospital, Harvard Medical School, Boston, MA, USA

Corresponding Authors: Fan Zhang (zhangfanmark@gmail.com) and Lauren J. O'Donnell (odonnell@bwh.harvard.edu)

**Abstract**

The relationship between brain connections and non-imaging phenotypes is increasingly studied using deep neural networks. However, the local and global properties of the brain's white matter networks are often overlooked in convolutional network design. We introduce TractGraphFormer, a hybrid Graph CNN-Transformer deep learning framework tailored for diffusion MRI tractography. This model leverages local anatomical characteristics and global feature dependencies of white matter structures. The Graph CNN module captures white matter geometry and grey matter connectivity to aggregate local features from anatomically similar white matter connections, while the Transformer module uses self-attention to enhance global information learning. Additionally, TractGraphFormer includes an attention module for interpreting predictive white matter connections. We apply TractGraphFormer to tasks of sex and age prediction. TractGraphFormer shows strong performance in large datasets of children (n=9345) and young adults (n=1065). Overall, our approach suggests that widespread connections in the WM are predictive of the sex and age of an individual. For each prediction task, consistent predictive anatomical tracts are identified across the two datasets. The proposed approach highlights the potential of integrating local anatomical information and global feature dependencies to improve prediction performance in machine learning with diffusion MRI tractography.

# 1 Introduction

The brain's white matter (WM) connections (fiber tracts) and their tissue microstructure can be quantitatively mapped using diffusion magnetic resonance imaging (dMRI) tractography (Zhang et al., 2022b). The white matter fiber tract connections have important inter-individual variability, with implications for understanding neurodevelopment and disease (Forkel et al., 2022). Accordingly, one recent avenue of research explores white matter variability by predicting non-imaging phenotypes from dMRI tractography data using machine learning methods (Y. Chen et al., 2024; Gong et al., 2021; He et al., 2022; Xue et al., 2024). Non-imaging phenotypes can include demographic, behavioral, clinical and cognitive measures of subjects. These non-imaging phenotypes are predicted from input features describing the brain's white matter connections, such as connectivity "strengths" or tissue microstructure measures (Zhang et al., 2022b).

The prediction tasks have been studied using traditional machine learning models such as Support Vector Machines (SVMs) (Feng et al., 2022; Zhang et al., 2018a), as well as deep methods such as convolutional neural networks (CNNs) (Chen et al., 2020; He et al., 2022; Jeong et al., 2021; W. Liu et al., 2023; Lo et al., 2024b; Wei et al., 2023) or point-based neural networks (Chen et al., 2022; Y. Chen et al., 2024). For example, (Feng et al., 2022) evaluated machine learning methods using multiple features describing WM connections for the prediction of individual performance across cognitive domains (Feng et al., 2022). In another example, (He et al., 2022) performed age and sex prediction using a CNN with input white matter microstructure and connectivity features extracted from dMRI tractography data.

However, challenges remain in the design of deep neural networks for prediction tasks based on white matter features. The brain's white matter is organized into structural networks that have important local and global properties (Park and Friston, 2013). Thus we investigate two main deep network design challenges related to the local and global properties of WM features.

First, it is a challenge to leverage existing neuroanatomical knowledge about local neighborhoods of WM connections to guide neural network training. In this paper, we use the phrase "local neighborhoods" to refer to WM connections with similar anatomy. As illustrated in Fig. 1, we define similar WM connections as those with similar WM geometry (trajectory and location) or gray matter (GM) connectivity (connected GM regions). By guiding network convolutions to focus on local neighborhoods of connections that share similar anatomy, local features can be learned from these neighborhood connections, and we hypothesize that local feature learning and overall network performance can be enhanced. Though prediction tasks can potentially benefit from useful neuroanatomical information about the brain's WM connections, off-the-shelf methods such as SVMs or CNNs generally ignore all anatomical information about the white matter connections.

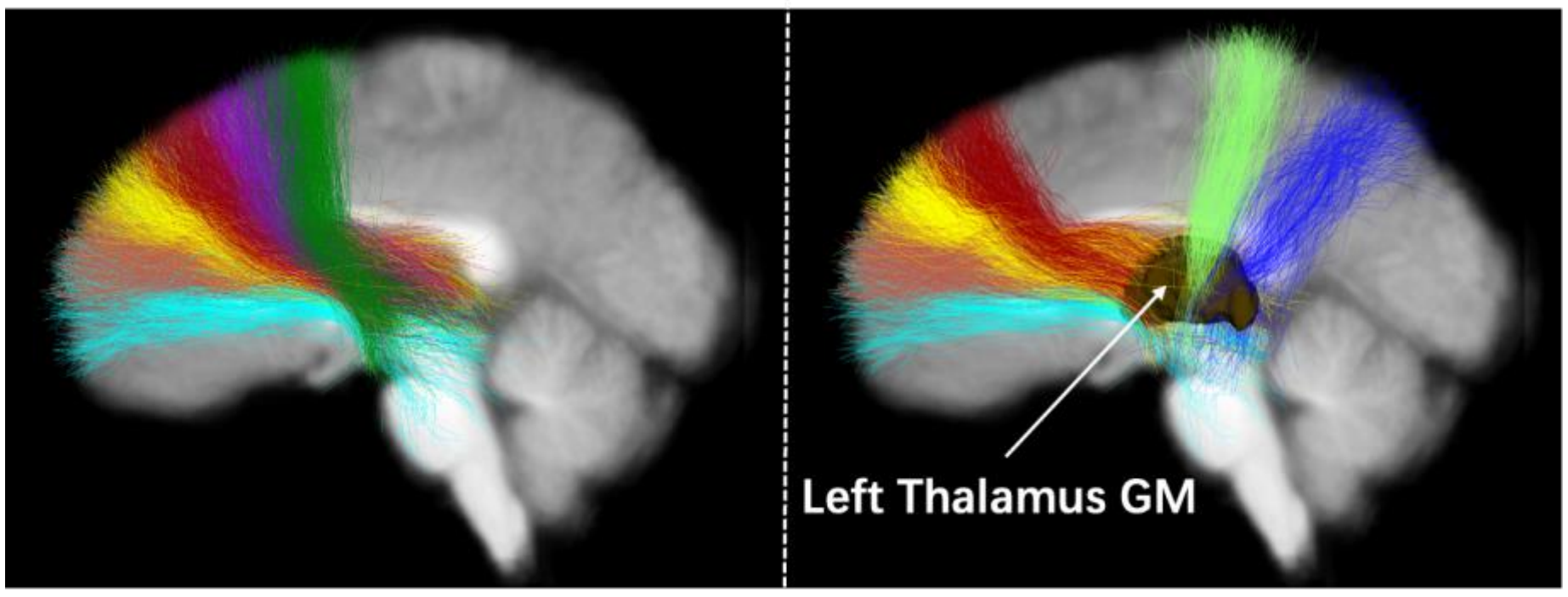

Fig. 1. Illustrations of WM connections with similar WM geometry (left) or GM connectivity (right). Each connection, or white matter fiber cluster, is shown in a unique color.

Second, it is a challenge for neural networks to capture the global information of WM connections. In this paper, we consider "global WM information" to be the information from all WM connections within a brain and the global relationships between these WM connections. Such global WM information is expected to be important for tractography-related prediction tasks because the human brain includes multiple interconnected structural networks that work together. For example, incorporating global information from randomly selected whole-brain streamlines improved streamline classification performance for tractography parcellation compared to using only local information from a single streamline (Xue et al., 2023). We hypothesize that enhancing the learning of global information will improve network performance for tractography-based prediction. However, it is generally a challenge for off-the-shelf methods such as CNNs to fully capture global information (A. He et al., 2023).

To address the first challenge, a few studies have aimed to develop dedicated neural networks to utilize anatomical information for analyses of the brain's structural connections. The BrainNetCNN (Kawahara et al., 2017) includes novel convolutional filters that improve performance (Yeung et al., 2020) by handling the topology of connectivity matrices (where each row or column corresponds to a gray matter (GM) region or node, and entries or edges in the matrix indicate connectivity strengths derived from the tractography connecting the GM regions). Other approaches apply graph convolutional neural networks to connectivity matrices, e.g., (Ktena et al., 2018). However, the above classes of methods are restricted to the anatomical information contained in the row and column structure of the connectivity matrix, and they cannot leverage any additional anatomical information to inform network convolutions.

To address the second challenge, a few studies have investigated the application of Transformers, which can learn global information and capture long-range interactions (A. He et al., 2023) in natural language processing (Vaswani et al., 2017) and computer vision tasks (Dosovitskiy et al., 2021). The initial applications of Transformers to tractography include one study that demonstrated benefits of learning global connectome patterns for the classification of Parkinson's disease (Machado-Reyes et al., 2022). Another study (Zhang et al., 2022d) demonstrated the potential of transformers to identify discriminative tractography streamlines in the classification of schizophrenia. The advantages of transformers relate to their network structure. Comprising attention-driven building blocks, Transformers use a self-attention mechanism to model long-range interactions between all tokens (Vaswani et al., 2017). However, compared to CNNs, Transformers are weak in local feature learning and require larger training datasets.

Hybrid networks blending CNNs and Transformers have been proposed to leverage the strengths of each: CNNs for local feature learning and Transformers for global long-range dependency learning. These hybrids have found success in many computer vision applications (Wu et al., 2021; Zhang et al., 2023; Zheng et al., 2022). One related tractography analysis application has explored a hybrid network (Machado-Reyes et al., 2022) to successfully classify patients with Parkinson's disease based on the structural connectome of the brain. This study proposed a basic Transformer architecture that leveraged a CNN for encoding the input connectomes. In our approach, we propose a hybrid CNN-Transformer network that is designed to specifically exploit the unique characteristics of white matter structures in dMRI tractography.

To focus our project, we choose two testbed problems of sex and age prediction. Sex differences in the brain have important implications for neurobehavioral research (Choleris et al., 2018) and understanding the sex-specific clinical trajectories of brain disease and aging (Cosgrove et al., 2007; McCarthy, 2016; Ruigrok et al., 2014; Walsh et al., 2021). Sex is an important source of WM variability (Ingalhalikar et al., 2014), and the WM has both local and global sex-related differences in its tissue microstructure (J. Chen et al., 2024) that evolve over the human lifespan (Conte et al., 2024). As a testbed task, sex prediction has the potential to provide clinically or neuroscientifically useful information about the brain. Multiple studies have shown that the prediction of sex from neuroimaging data may shed light on sex-based differences in brain structure and function and how they evolve over the lifespan (Cosgrove et al., 2007; Kurth et al., 2021; Weis et al., 2020). While sex prediction is not a straightforward problem, (Eikenes et al., 2022; Sanchis-Segura et al., 2022), several studies have demonstrated that features from quantitative dMRI tractography (Zhang et al., 2022c) are useful for sex classification (He et al., 2022; Kulkarni et al., 2013; Yeung et al., 2023a, 2020). Age is another important source of WM variability and strongly affects WM microstructure in the brain, which changes over the human lifespan as the brain undergoes neurodevelopment and aging (Lebel et al., 2012). Age prediction from neuroimaging data allows estimation of an individual's "brain age" and their "brain-age gap," the difference between predicted brain age and actual chronological age, as described in a recent review (Soumya Kumari and Sundarrajan, 2024). Brain age prediction is an important testbed task because it has been shown to produce clinically relevant biomarkers related to mortality (Cole et al., 2018), cognitive impairment (Lee et al., 2022), and early identification of brain-based illnesses (Soumya Kumari and Sundarrajan, 2024). Several studies have demonstrated that features from quantitative dMRI tractography (Zhang et al., 2022c) are useful for brain age prediction (Han et al., 2014; He et al., 2022; Yeung et al., 2023a). However, most existing methods for sex and age prediction have used a structural connectome matrix as input, which provides limited information about brain anatomy (Han et al., 2014; Kulkarni et al., 2013; Lin et al., 2016; Yeung et al., 2023a, 2020). In comparison with the traditional structural connectome matrix, fiber clusters enable consistent parcellation across the lifespan (Zhang et al., 2018c) with higher test-retest reproducibility (Zhang et al., 2019) and improve power to predict human traits (R. Liu et al., 2023). In addition, fiber clusters facilitate a compact vector representation of the connectome (R. Liu et al., 2023; Zhang et al., 2018a) and enable a variety of downstream statistical and machine learning analyses (Chen et al., 2023b; Gabusi et al., 2024; He et al., 2022; R. Liu et al., 2023; Xue et al., 2024; Zhang et al., 2018b). In this work, we leverage features from a white matter tractography fiber cluster parcellation (Zhang et al., 2018c), which provides WM geometry and GM anatomical information for each cluster (Fig. 1) that we employ to enhance network design.

In this work, we propose a deep learning framework, TractGraphFormer, that focuses on local white matter anatomical neighborhoods while learning about global long-range feature dependencies among the brain's connections. We design a hybrid two-stream neural network consisting of a Graph CNN and a Transformer module for sex and age prediction. The designed Graph CNN stream incorporates information about anatomical neighborhoods of WM connections to enhance performance of CNNs. The Transformer stream is integrated into the framework in parallel to the Graph CNN to establish long-range dependencies among clusters by self-attention, providing a supplement to the local convolution in the Graph CNN stream. A feature-level fusion strategy is employed to aggregate the local and global information learned by the Graph CNN and Transformer streams. In addition, an attention module is adopted to provide interpretability about the predictive WM tracts for sex classification and age prediction. The main contributions of this paper are summarized as follows:

- We propose TractGraphFormer, a hybrid network consisting of Graph CNN and Transformer feature extractors for sex and age prediction based on fiber cluster features.
- To learn anatomically informed local features, we model the anatomical relationship between clusters as a graph informed by WM geometry and/or GM connectivity. Then features from anatomically similar clusters are aggregated using a Graph CNN to improve performance.
- The Transformer stream is designed to extract global features and build long-range dependencies from all input clusters by naturally taking each cluster as a token.
- We propose to integrate a gated attention module into our framework to enable interpretation of predictive fiber clusters.
- We evaluate our method on two large-scale datasets of children and healthy young adults. The results show that the proposed method outperforms state-of-the-art methods in sex and age prediction performance and identifies consistent predictive white matter connections across datasets.

The current paper extends a preliminary version of the work (Chen et al., 2023b) where we proposed the TractGraphCNN framework. In the current paper, we incorporate several technical improvements and additional analyses. First, based on the previous TractGraphCNN framework which extracts local features, we extend the network by adding a parallel Transformer module to extract global features and then fuse them to improve prediction performance. We call the new proposed framework TractGraphFormer. Second, we propose a new combined graph that extends the two types of graphs in the preliminary work by simultaneously considering WM and GM information. We demonstrate the effectiveness of our method on the three types of graphs. Third, we conduct an additional age prediction task to further evaluate our proposed TractGraphFormer. The experiment results show that TractGraphFormer has superior age prediction performance and enables identification of predictive white matter tracts for age prediction. We also include related discussions in the paper. Finally, we perform an extended set of experiments, including parameter tuning experiments to investigate prediction performance under different parameter values and ablation studies to evaluate the effectiveness of our proposed modules. By incorporating these improvements, the paper strengthens the overall methodology and provides a more comprehensive evaluation of the proposed TractGraphFormer framework.

## 2 Methods

The overall workflow of the proposed method includes three steps. First, WM features are extracted from dMRI tractography fiber clusters (Section 2.1). Second, a graph is built to model the anatomical relationship between fiber clusters, informed by WM geometry and/or GM connectivity information (Section 2.2). Third, the built graph and WM features are input to the proposed TractGraphFormer framework (Section 2.3) for sex classification and age prediction and interpretation of predictive tracts.

### 2.1 Study Dataset and Feature Extraction

In this study, we used two datasets of quantitative cluster-specific WM features. dMRI acquisitions were obtained from the Adolescent Brain Cognitive Development (ABCD) (Casey et al., 2018) and Human Connectome Project Young Adult (HCP-YA) (Van Essen et al., 2013) studies, and dMRI processing was performed as described below.

#### 2.1.1 ABCD Dataset

To evaluate the performance of our proposed method, this study utilized dMRI data and demographic data (sex and age) of 9345 young children (47.8% female, 9.9 +/- 0.6 years old) from the large-scale, multi-site ABCD dataset (Casey et al., 2018). The dMRI data was originally acquired and minimally preprocessed by the ABCD study Data Analysis, Informatics, and Resource Center group (b = 500, 1000, 2000 and 3000 s/mm$^2$, resolution = 1.7 × 1.7 × 1.7 mm$^3$) (Zhang et al., 2022a). In this study, we used a derived database of harmonized and quality-controlled dMRI (Cetin-Karayumak et al., 2024) that was harmonized across 21 acquisition sites to remove scanner-specific biases while preserving inter-subject biological variability (Xue et al., 2024; Zhang et al., 2022a). For each subject, the dMRI data from the b = 3000 shell of all gradient directions and all b = 0 scans were extracted (Casey et al., 2018; Cetin-Karayumak et al., 2024). These data were used to perform tractography because this single shell can reduce computation time and memory usage while providing the highest angular resolution for tractography (Descoteaux et al., 2007; Ning et al., 2015; Zekelman et al., 2022).

#### 2.1.2 HCP-YA Dataset

We also conducted experiments on the dMRI and demographic data (sex and age) of 1065 healthy young adults (54.0% female, 28.7 +/- 3.7 years old) from the minimally preprocessed Human Connectome Project Young Adult (HCP-YA) dataset (b = 1000, 2000 and 3000 s/mm2, TE/TR =89/5520 ms, resolution = 1.25 × 1.25 × 1.25 mm3) (Van Essen et al., 2013). For each subject, the dMRI data from the b = 3000 shell of all 90 gradient directions and all 18 b = 0 scans were extracted (Descoteaux et al., 2007; Ning et al., 2015; Zekelman et al., 2022).

#### 2.1.3 White Matter Fiber Cluster Features

Whole brain tractography and white matter feature extraction were performed for the ABCD and HCP-YA datasets as follows (Cetin-Karayumak et al., 2024; Zekelman et al., 2022). First, whole brain tractography was obtained from the dMRI data using a two-tensor unscented Kalman filter tractography method (UKFt) (Reddy and Rathi, 2016) via SlicerDMRI (Norton et al., 2017; Zhang et al., 2020), as implemented in the ukftractography package[1]. UKFt is highly consistent across the human lifespan, test-retest scans, disease states, and different acquisitions (Zhang et al., 2018c), and outperforms state-of-the-art tractography methods for reconstructing anatomical somatotopy (J. He et al., 2023). In contrast to other tractography methods that fit a model to the diffusion signal independently at each voxel, the UKFt framework employs prior information from the previous step during each tracking step to help stabilize model fitting (Zhang et al., 2018c). The two-tensor model adopted in the UKFt tractography algorithm can depict crossing streamlines, which are prevalent in white matter (Farquharson et al., 2013; Vos et al., 2013).

Second, tractography was parcellated using a machine learning approach that has been shown to consistently identify WM tracts across datasets, acquisitions, and the human lifespan (Zhang et al., 2018c), as implemented in the WMA package[2]. This approach relies on an anatomically curated tractography atlas, the O'Donnell Research Group (ORG) Atlas (Zhang et al., 2018c). The ORG atlas was generated by creating dense tractography maps of 100 individual subjects and then applying a fiber clustering method (Lauren J. O'Donnell and Westin, 2007) to group the streamlines across subjects according to their similarity in shape and location. The resulting atlas fiber clusters, which describe common brain connections and their variability in the population, can be robustly identified across the human lifespan (Zhang et al., 2018c). The ORG atlas fiber clusters were viewed and annotated by an expert neuroanatomist, who categorized the fiber clusters to define WM tracts based on neuroanatomical expertise (Zhang et al., 2018c). A description of the WM geometry and GM connectivity of each fiber cluster is provided in the ORG atlas, which includes 953 expert-curated fiber clusters categorized into 75 WM tracts. The ORG atlas enables automated exclusion of false positive streamlines. For each subject, parcellation produced 953 subject-specific fiber clusters, where each subject-specific fiber cluster corresponded to a specific atlas fiber cluster. Importantly, cluster IDs were assigned according to the atlas and corresponded across subjects (e.g., cluster #1 corresponded across all subjects and datasets studied).

Finally, quantitative white matter features were computed for all clusters in both datasets. As in previous studies of ABCD and HCP (Cetin-Karayumak et al., 2024; Zekelman et al., 2022), we used three computed WM features, the mean fiber cluster fractional anisotropy (FA), mean fiber cluster mean diffusivity (MD), and number of streamlines (NoS). FA and MD are fiber-specific microstructure measures computed from the multi-tensor model employed to perform tractography. These measures quantify the anisotropy and magnitude of water diffusion, providing indirect measures of the condition and geometry of the white matter tissue. These measures were computed following the typical practice of averaging across the microstructure values of all streamline points in the fiber cluster (Zhang et al., 2022b). As a popular measure for the analysis of tractography, NoS was also computed to indicate the connectivity strength or connection "strength" of clusters (Zhang et al., 2022b). The above measures (FA, MD, and NoS) have previously been shown to be informative for the sex and age prediction tasks (J. Chen et al., 2024; Yeung et al., 2023b, 2020). Since NoS is affected by intracranial volume, a common confound in sex prediction (Sanchis-Segura et al., 2022), in this work we propose to normalize NoS to reduce the effect of brain size by using the percentage of streamlines (PoS), which was calculated as the number of streamlines of a cluster divided by the total number of streamlines in all clusters of the subject. The input data for each subject (in ABCD and HCP-YA) was a cluster-wise WM feature matrix of size 3 x 953. For absent clusters due to individual anatomical variation, we set features to zero (He et al., 2022; Xue et al., 2024). Finally, a min-max normalization was performed on the input feature matrix for FA, MD, and PoS individually.

## 2.2 Anatomically Informed Graph Construction

We propose to build graphs that model the anatomical relationship between fiber clusters. Each cluster is represented as a node in the proposed graph, with its cluster-specific WM features as node features. Edges are built to connect neighboring fiber clusters with similar anatomy. We propose three types of anatomically informed graphs, as described below.

[1] https://github.com/pnlbwh/ukftractography.

[2] https://github.com/SlicerDMRI/whitematteranalysis.

### 2.2.1 Fiber Tract Geometry Informed Graph

The first type of graph (WMG) proposed in our study is based on WM tractography fiber geometric similarity, a well-established concept in the field of fiber clustering (Zhang et al., 2022b). The neighborhoods in WMG are defined according to fiber tract geometry. Specifically, we first compute the geometric distance between each pair of fiber clusters in the ORG atlas, which is measured as the mean of the pairwise fiber distances between the streamlines in the two fiber clusters. We use a popular fiber distance measure that has been used for over 20 years in the fiber clustering field, the mean closest point (computed using the minimum direct flip method) fiber distance (Garyfallidis et al., 2018; L. J. O'Donnell and Westin, 2007). A low distance between two clusters represents a high similarity in terms of WM anatomy. Then, for each cluster, we choose the top $k$ clusters with the lowest geometric distances as neighbors, and edges are placed in between for graph construction.

### 2.2.2 Cortical and Subcortical Connectivity Informed Graph

The second type of graph (GMG) proposed in our study is based on GM regions to which the fiber clusters connect. The GMG neighborhoods are strategically defined based on the GM regions connected by fiber clusters, leveraging atlas population-based anatomical characteristics provided in the ORG atlas. Fiber clusters may intersect multiple GM regions. To ensure a robust representation of connectivity, we focus on the top two GM regions with the highest intersection for each fiber cluster, capturing the most relevant connections. For each cluster, we identify its connected FreeSurfer (Fischl, 2012) GM regions using the ORG atlas, which provides the percentage of streamlines intersecting each region. We leverage this information to identify the top two FreeSurfer regions most commonly intersected by the streamlines of each cluster. The neighborhood of a cluster is then defined as the set of clusters with these top two FreeSurfer regions in common, and edges are placed in between for graph construction.

### 2.2.3 Combined WM and GM Informed Graph

A third type of graph, namely the combination of WMG and GMG (CMG), is proposed in this work to simultaneously utilize WM and GM information for modeling anatomical relationships between clusters. Specifically, for each cluster (node) in CMG, its neighborhood clusters (nodes) are defined as clusters that both belong to the top $k$ clusters with the lowest geometric distances and have the top two FreeSurfer regions in common.

### 2.3 Network Architecture

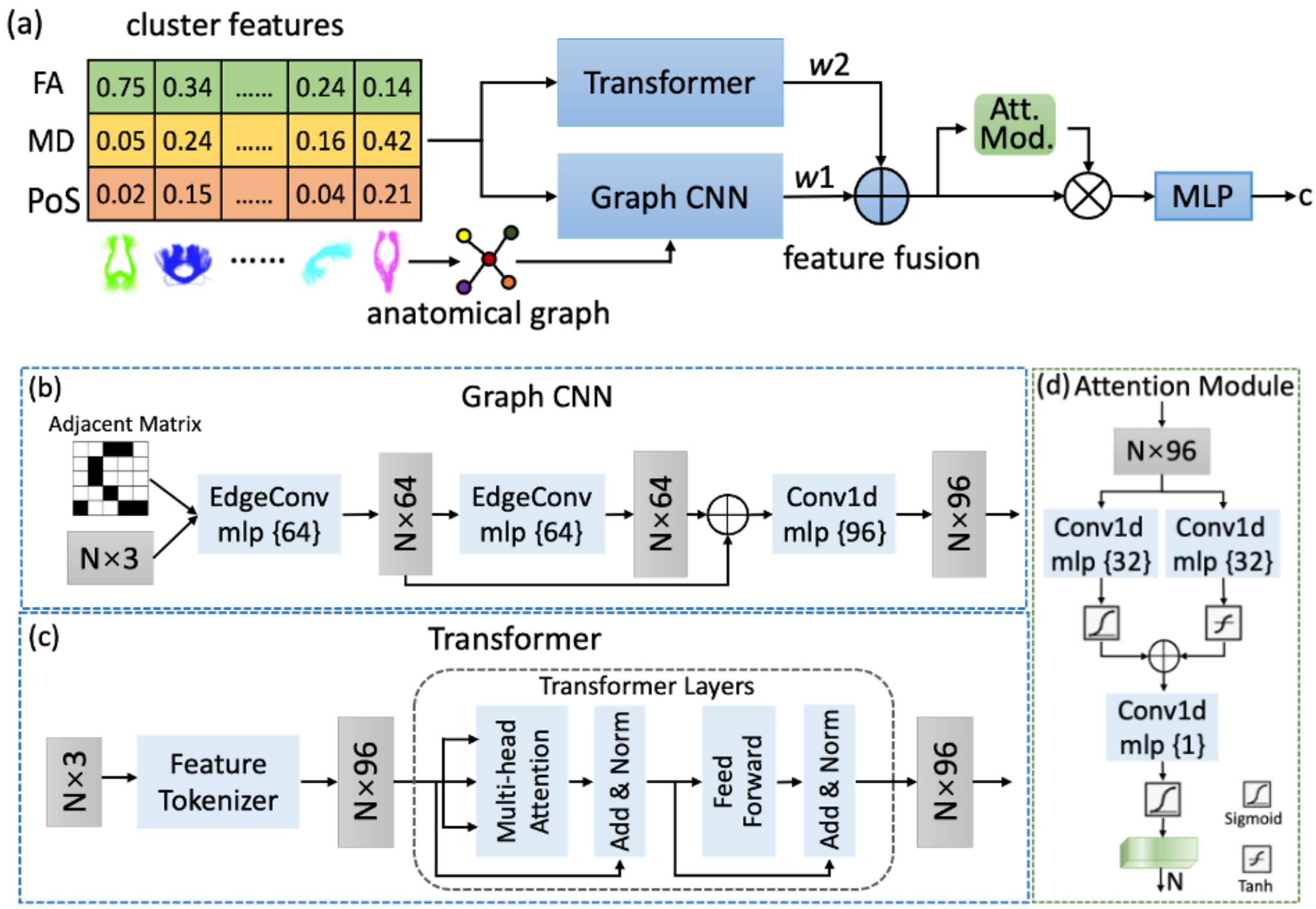

Fig. 2. (a) Overall pipeline of TractGraphFormer. (b) Network structure of the Graph CNN stream. (c) Network structure of the Transformer stream. (d) Network structure of the attention module. (N: number of clusters)

#### 2.3.1 Overall Pipeline

The overall architecture of our proposed TractGraphFormer framework is shown in Fig. 2 (a). The proposed framework is a hybrid two-stream deep neural network incorporating Graph CNN and Transformer modules for feature extraction. Local and global features are extracted from the Graph CNN and the Transformer, respectively, and then these features are fused in a weighted sum manner. The Graph CNN stream extends the 1D-CNN model (He et al., 2022) for group classification using fiber cluster features by replacing the 1D convolutional layers in the original model with EdgeConv layers (Wang et al., 2019) to utilize the information of anatomically neighboring clusters (Section 2.3.2). We apply a Transformer (Gorishniy et al., 2021; Vaswani et al., 2017) with a self-attention mechanism that can naturally take each cluster as a token to learn long-range dependencies among clusters (Section 2.3.3). After fusion of local and global features, we add a gated attention module (Ilse et al., 2018) in the network that can assess the importance of each cluster to enable interpretation of results (Section 2.3.4). Then a flatten operation and two fully-connected layers follow to obtain the final classification results.

#### 2.3.2 Graph CNN Stream

In the Graph CNN stream, we adopt EdgeConv to aggregate information from neighboring nodes that represent anatomically similar fiber clusters (Fig. 3). EdgeConv was proposed in the popular Dynamic Graph CNN (DGCNN) method to capture the local geometric structure of point clouds (Wang et al., 2019). The basic idea of EdgeConv is to use a learnable fully-connected layer to compute an edge feature of two neighboring nodes $x_i$ and $x_j$ based on their input features. Then, EdgeConv outputs the refined node features by aggregating the learned edge features with max-pooling. This learning process enables dynamic updates of graph structure by recomputing distances of points in the feature space. In our application, as the proposed graphs are built to explicitly reflect the anatomical relationship between fiber clusters, we maintain a static graph structure across layers by using the same graph structure across all EdgeConv layers without recomputing distances between node features.

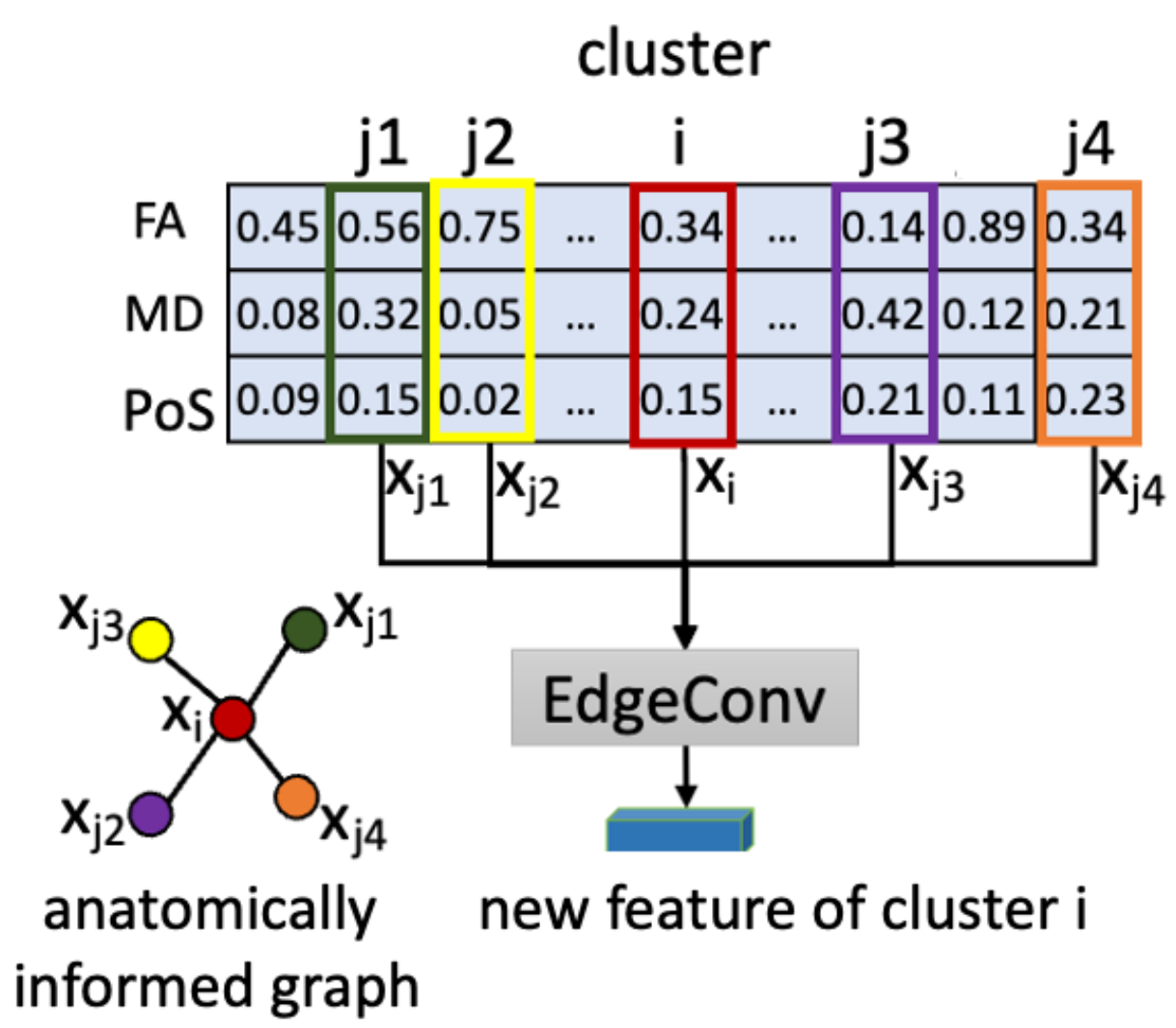

Fig. 3. Graphic illustration of the usage of EdgeConv to leverage fiber cluster neighborhood information (left), with comparison to the standard 1-D convolutional layer (right). In EdgeConv, new features of a target cluster are learned from its neighboring clusters defined by the anatomically informed graph (illustrated in the bottom-left corner of the left figure), whereas in 1D Conv, features are learned based on neighboring clusters defined by their cluster IDs.

The overall architecture of the Graph CNN stream is shown in Fig. 2 (b). The inputs of Graph CNN are the constructed graphs (Section 2.2) including node (cluster) features and adjacency matrices defining edges between nodes. We use two EdgeConv layers and one 1-D convolutional layer to extract features. The two EdgeConv layers compute edge features with two fully-connected layers. Shortcut connections are included to extract multi-scale features and one 1-D convolutional layer (kernel size=1) is followed to aggregate multi-scale features obtained from the previous EdgeConv layers. Unlike general graphs, the nodes (clusters) in our constructed graph are arranged in order and correspond across subjects. Traditional Graph CNNs use max pooling, which disregards node order and would cause a loss of information about cluster ordering and correspondence across subjects. Therefore, after feature extraction, we do not use the max pooling operation as in traditional Graph CNNs, but instead, we retain the flatten operation in the 1D-CNN model (He et al., 2022) to preserve the information about cluster correspondence across subjects. The output of the Graph CNN stream is a feature matrix, where one dimension corresponds to clusters and the other dimension to the learned feature vector of each cluster.

### 2.3.3 Transformer Stream

Fig. 2 (c) illustrates the usage of the Transformer stream to capture long-range dependencies among cluster features. Specifically, the Transformer module used for feature extraction is adapted from the model in (Gorishniy et al., 2021), which is a simple adaptation of the vanilla Transformer architecture (Vaswani et al., 2017) for 1-D tabular data. As in (Gorishniy et al., 2021), a Transformer block comprises two parts, a feature tokenizer module and stacked Transformer layers, consisting of a self-attention and a feedforward module. For our input data, we naturally take each cluster as a token and we get three features (FA, MD, and NoS) for each input token. Then, the feature tokenizer module transforms the input features into embeddings with a specified dimension. The embeddings are then input to the stacked Transformer layers to learn token features from all tokens. In each Transformer layer, the input features are first linearly mapped as query (Q), key (K), and value (V) , which are the keys of the multi-head attention mechanism. The process is shown below (Vaswani et al., 2017):

$$[q, k, v] = xW_{qkv}, W_{qkv} \in R^{n \times d_{pkv}} \quad (1)$$

where $n$ is the dimension of input embeddings for each token and $d_{qkv}$ is the dimension of query/key/value. Then the output of the self-attention layer is computed as a weighted sum of the values, where the weight assigned to each value is computed by a compatibility function of the query with the corresponding key. The output attention map is calculated as (Vaswani et al., 2017):

$$Attention(Q, K, V) = softmax(\frac{QK^T}{\sqrt{d_k}})V \quad (2)$$

where $d_k$ is the dimension of the key (Vaswani et al., 2017). Following that, a fully connected feed-forward network is applied to the obtained features of each token from the self-attention layer and generates the output features of the Transformer layer. The "classification token" in (Gorishniy et al., 2021) is removed from our model because the goal of the Transformer module is to extract global features for each token instead of obtaining a final classification score. In this way, the output of the Transformer module is a feature matrix, where one dimension corresponds to clusters and the other dimension corresponds to the learned feature vector of each cluster.

#### 2.3.4 Fusion of Features from the Two-stream Model

After feature extraction from the Graph CNN and Transformer, respectively, we adopt a feature-level fusion strategy using a weighted sum to aggregate features from the two streams. Specifically, two learnable weights ($w1$ and $w2$) are assigned to the two streams and the fused features are calculated as the weighted sum of the features extracted from the Graph CNN and Transformer streams. The weights of the two streams are initialized and updated during model training to get the best classification performance.

#### 2.3.5 Attention Module and Interpretation

For the purpose of interpretation, it is important to identify highly predictive WM connections for the tasks of sex classification and age prediction. To achieve this, we add an attention mechanism after feature fusion using the popular gated attention module from (Ilse et al., 2018). The attention module (Fig. 2(d)) is composed of two parallel fully-connected layers followed by a sigmoid and tanh activation functions, a concatenation operation, and another fully-connected layer followed by a sigmoid function. The output is a 1-D attention map of size 953 with values between 0 and 1, indicating the importance of the corresponding cluster to the classification task. Thus at test time an attention score is obtained for each cluster in each subject.

Next, we identify the most predictive anatomical WM tracts. We first compute the mean score of each cluster across all testing subjects. Then we identify the top 50 clusters with the highest importance scores as highly predictive clusters for each prediction task. Finally, we identify all tracts to which these predictive clusters belong, according to the ORG atlas (Zhang et al., 2018c).

#### 2.3.6 Implementation Details

We adopt 5-fold cross-validation for model training and performance evaluation. In each fold, 7473 subjects (80%) of ABCD subjects are used for training the neural network, while 1869 (20%) are used for testing. For the HCP subjects, 852 subjects (80%) are used for training, while 213 (20%) are used for testing. The averaged prediction performance of all folds and standard deviation across folds are reported in the experimental results. We train and test each classification model on each dataset independently.

All experiments are performed on an NVIDIA RTX A4000 GPU using Pytorch (v1.12.1) (Paszke et al., 2019). Our overall network is trained for 100 epochs with a learning rate of 1e-3. The batch size of training is 32 and Adam (Kingma and Ba, 2014) is used for optimization. The number of selected WM-informed neighbors $k$ is set to 30, which obtains the best performance in the parameter tuning experiment (see Section 3.1). For the Transformer module, the number of layers and number of heads are set to 1 considering the relatively small training dataset of medical images. The token dimension is 96, as well as the dimension of output features from the Graph CNN and Transformer streams. This number is chosen to balance between classification performance and computation cost. The weights of the two streams (*w*1 and *w*2) are initialized with 0.5.

## 3 Experiments and Results

We evaluate our proposed TractGraphFormer on two testbed tasks of sex classification and age prediction on both ABCD and HCP datasets. First, we perform a parameter tuning experiment (Section 3.1). Second, we compare our proposed method with four state-of-the-art methods (Section 3.2). Third, we perform ablation studies to evaluate the effectiveness of each component of TractGraphFormer (Section 3.3). Finally, weights of Graph CNN and Transformer streams and highly predictive tracts for sex classification or age prediction are identified for interpretation (Section 3.4) for both datasets.

Four evaluation metrics are adopted in our study to evaluate sex classification performance: accuracy, precision, recall and F1 score. For precision, recall, and F1 score, the averaged values of the two classes are calculated for evaluation. For age prediction, we adopt the evaluation metric of mean absolute error (MAE), which has been commonly used in previous studies of brain age prediction (He et al., 2022; Yeung et al., 2023a).

### 3.1 Parameter Tuning Experiment

We investigated the influence of the number of white-matter-geometry-defined neighbors (*k*) on sex classification performance. We focus on CMG graph construction (see Section 2.2.3 for details). We explored the sex classification accuracies under five *k* values (10, 20, 30, 40, 50) for both datasets, as shown in Table 1. While the model performance is relatively robust to the choice of *k*, the performance of our model initially increases with the increase of *k* values, achieves a peak at *k*=30, and then decreases. This is likely because the aggregation of information from more anatomical neighbors benefits the performance gradually at first, but then the influence of distant clusters might mitigate the performance as the number of neighbors (*k*) increases.

Table 1 Accuracy across different numbers of neighbors (*k*) during graph construction for both datasets.

| *k* | 10 | 20 | 30 | 40 | 50 |
|---|---|---|---|---|---|
| ABCD | 85.50 (1.22) | 85.81 (0.90) | **86.25 (0.83)** | 85.99 (0.77) | 86.01 (0.89) |
| HCP | 94.46 (1.55) | 94.84 (2.26) | **95.59 (2.17)** | 95.31 (1.71) | 94.84 (1.83) |

### 3.2 State-of-the-art (SOTA) Comparison Experiments

For both sex and age prediction tasks, we compared the performance of our proposed method with four SOTA methods: SVM, 1D-CNN (He et al., 2022), $\text{TractGraphCNN}_{\text{WMG}}$, and $\text{TractGraphCNN}_{\text{GMG}}$. SVM is one of the most commonly used machine learning algorithms for tractography-related classification tasks (Deng et al., 2019; Kim and Park, 2016; Zhang et al., 2018a). The 1D-CNN model in (He et al., 2022) was specifically proposed for sex classification and age prediction based on cluster-wise white matter features, and has demonstrated competitive performance. $\text{TractGraphCNN}_{\text{WMG}}$ and $\text{TractGraphCNN}_{\text{GMG}}$ are our preliminary work (Chen et al., 2023b) with two graph types, WMG and GMG, respectively. For the 1D-CNN, $\text{TractGraphCNN}_{\text{WMG}}$ and $\text{TractGraphCNN}_{\text{GMG}}$ models, we used the same parameter settings as in the original papers (Chen et al., 2023b; He et al., 2022). The sex classification results of all methods from the ABCD and HCP datasets are shown in Tables 2 and 3, respectively, and the age prediction results are shown in Table 4. To conduct statistical analysis on the results of evaluation metrics, the one-way repeated measures Analysis of Variance (ANOVA) was adopted because it is a standard statistical method for comparing the performance of multiple machine learning models (Demšar, 2006). ANOVA was applied to the results of each evaluation metric, followed by post-hoc pairwise comparisons using paired t-tests between our proposed method and each comparison method.

Table 2 Comparison of sex classification performance across methods and graph types for ABCD dataset.

| Methods | Acc | Pre | Rec | F1 |
|---|---|---|---|---|
| SVM | 78.57 (2.07)*** | 78.61 (2.08)*** | 78.57 (2.00)*** | 78.52 (2.05)*** |
| 1D-CNN | 84.22 (1.00)** | 84.24 (1.05)** | 84.14 (0.94)** | 84.16 (0.98)** |
| $TractGraphCNN_{WMG}$ | 85.02 (1.32)* | 85.04 (1.30)* | 85.00 (1.23)* | 84.98 (1.29)* |
| $TractGraphCNN_{GMG}$ | 85.17 (1.27)* | 85.31 (1.54)* | 85.05 (1.13)* | 85.11 (1.21) |
| $TractGraphFormer_{WMG}$ | 86.09 (1.12) | 86.16 (1.22) | 86.03 (1.00) | 86.05 (1.08) |
| $TractGraphFormer_{GMG}$ | 85.97 (1.12) | 86.02 (1.03) | 85.98 (1.04) | 85.94 (1.10) |
| $TractGraphFormer_{CMG}$ | **86.25 (0.83)** | **86.29 (0.80)** | **86.21 (0.78)** | **86.21 (0.81)** |

p<0.0005*** p<0.005** p<0.05 *

Table 3 Comparison of sex classification performance across methods and graph types for HCP dataset.

| Methods | Acc | Pre | Rec | F1 |
|---|---|---|---|---|
| SVM | 90.61 (2.83)*** | 90.52 (2.82)*** | 90.64 (2.91)*** | 90.55 (2.85)*** |
| 1D-CNN | 92.58 (1.91)** | 92.55 (1.90)** | 92.59 (1.99)*** | 92.53 (1.93)** |
| $TractGraphCNN_{WMG}$ | 94.37 (2.75)* | 94.34 (2.78)* | 94.40 (2.70)* | 94.33 (2.76)* |
| $TractGraphCNN_{GMG}$ | 94.27 (2.29)* | 94.31 (2.22)* | 94.31 (2.27)* | 94.24 (2.29)* |
| $TractGraphFormer_{WMG}$ | 95.21 (1.96) | 95.27 (1.94) | 95.13 (1.98) | 95.17 (1.97) |
| $TractGraphFormer_{GMG}$ | 94.84 (1.33) | 94.79 (1.37) | 94.86 (1.27) | 94.80 (1.32) |
| $TractGraphFormer_{CMG}$ | **95.59 (2.17)** | **95.57 (2.22)** | **95.65 (2.10)** | **95.56 (2.17)** |

p<0.0005*** p<0.005** p<0.05 *

As shown in Tables 2 and 3, on the sex classification task our proposed TractGraphFormer outperforms all comparison methods across both datasets, indicated by the highest values of all four evaluation metrics on classification accuracy. The ANOVA analyses show significant differences in the evaluation metrics of classification performance of compared methods for sex classification in both ABCD and HCP datasets ($p<0.0001$ in all analyses). In addition, post-hoc paired t-tests show that $TractGraphFormer_{CMG}$ obtains significantly higher values than the four comparison SOTA methods for all evaluation metrics across both datasets ($p<0.05$ in all analyses, as indicated by asterisks in Tables 2 and 3).

Table 4 Comparison of age prediction performance (Mean Absolute Error) across methods and graph types for both datasets.

| Methods | ABCD (months) | HCP (years) |
|---|---|---|
| SVM | 6.51 (0.14)*** | 3.04 (0.08)** |
| 1D-CNN | 6.27 (0.23)*** | 2.96 (0.11)** |
| $TractGraphCNN_{WMG}$ | 6.07 (0.12)** | 2.87 (0.14)** |
| $TractGraphCNN_{GMG}$ | 6.09 (0.11)** | 2.86 (0.10)* |
| $TractGraphFormer_{WMG}$ | 5.80 (0.16) | 2.82 (0.12) |
| $TractGraphFormer_{GMG}$ | 5.92 (0.15)* | 2.80 (0.16) |
| $TractGraphFormer_{CMG}$ | **5.76 (0.19)** | **2.78 (0.12)** |

p<0.0005*** p<0.005** p<0.05*

As shown in Table 4, our proposed TractGraphFormer shows the best age prediction performance among all comparison methods across both datasets, indicated by the lowest values of the MAE metric. The ANOVA analyses show significant differences in the age prediction performance of compared methods in both ABCD and HCP datasets (p<0.0001 in all analyses). In addition, post-hoc paired t-tests show that TractGraphFormerCMG obtains significantly lower MAE values than the four comparison SOTA methods across both datasets (p<0.05 in all analyses, as indicated by asterisks in Table 4).

We also compared the sex and age prediction performance across the three proposed types of input graphs for the ABCD and HCP datasets, as shown in Tables 2 and 3 for sex classification and Table 4 for age prediction. The results indicate that for both prediction tasks, all graph types show competitive performance compared to SOTA methods, while CMG slightly outperforms WMG and GMG across both datasets. For sex classification, an ANOVA was performed among the evaluation metrics of the three graph types and discovered no significant difference. For age prediction, a significant improvement was only found in the comparison between the CMG and the GMG graph types for the ABCD dataset (p<0.05). These results show that our proposed framework achieves good performance across three types of graphs, indicating the effectiveness of our network design in improving sex classification and age prediction performance.

### 3.3 Ablation Studies

We performed ablation experiments to evaluate the contribution of various components to the overall TractGraphFormer framework. We investigated the sex classification and age prediction performance of four models: TractGraphFormer, TractGraphFormer without attention module (TractGraphFormer w/o Att.), Graph CNN, and Transformer. The Graph CNN is one stream of our TractGraphFormer framework, as described in Section 2.3.2. Compared to TractGraphCNN, the Graph CNN model does not include an attention module and the hyperparameters such as *k* are fine tuned to different values. The Transformer model is the other stream of TractGraphFormer, as described in Section 2.3.3, including traditional Transformer blocks. An ANOVA analysis was applied to each of the four evaluation metrics. Then post-hoc pairwise comparisons using paired t-tests were performed between the evaluation metrics of different models.

In the ablation study for the task of sex prediction, as shown in Tables 5 and 6, the evaluation metrics of TractGraphFormer are higher than all compared models. In the ablation study for the task of age prediction, as shown in Table 7, the MAE values of TractGraphFormer (with or without the attention module) are lower (better) than those of the Graph CNN and the Transformer models. The ANOVA analyses show significant differences in the evaluation metrics of the compared models for both tasks across both datasets ($p$<0.0001 in all analyses). The ablation study shows that the proposed TractGraphFormer architecture (with or without attention) significantly outperforms the basic Graph CNN or Transformer architectures for both sex and age prediction tasks (p<0.05 in all analyses, as indicated by asterisks in Tables 5, 6 and 7). Therefore, the combination of Graph CNN and Transformer (TractGraphFormer) significantly improves prediction performance compared to when only one stream is used for feature extraction. The addition of attention slightly improves sex prediction in both the ABCD and HCP datasets and age prediction in the ABCD dataset, though these improvements are not significant compared to the TractGraphFormer without attention. However, the added benefit of the attention module is that it can facilitate downstream interpretation. Overall, these ablation results demonstrate the effectiveness of our network design to combine local and global information learned from the Graph CNN and the Transformer, respectively, for prediction tasks.

Table 5 Ablation study of network components for sex classification on the ABCD dataset.

| Models | Acc | Pre | Rec | F1 |
|---|---|---|---|---|
| Graph CNN | 85.04 (1.08)** | 85.09 (1.00)** | 85.04 (0.98)** | 85.01 (1.05)** |
| Transformer | 85.43 (1.22)* | 85.46 (1.23)* | 85.37 (1.16)* | 85.38 (1.20)* |
| TractGraphFormer w/o Att. | 86.00 (0.92) | 86.00 (0.91) | 85.98 (0.90) | 85.96 (0.91) |
| TractGraphFormer | **86.25 (0.83)** | **86.29 (0.80)** | **86.21 (0.78)** | **86.21 (0.81)** |

$p<0.005$** $p<0.05$ *

Table 6 Ablation study of network components for sex classification on the HCP dataset.

| Models | Acc | Pre | Rec | F1 |
|---|---|---|---|---|
| Graph CNN | 94.65 (1.92)* | 94.65 (1.89)* | 94.57 (1.98)* | 94.60 (1.94)* |
| Transformer | 91.64 (1.61)*** | 91.85 (1.43)*** | 91.51 (1.51)*** | 91.56 (1.61)*** |
| TractGraphFormer w/o Att. | 95.40 (1.79) | 95.41 (1.87) | 95.35 (1.76) | 95.35 (1.79) |
| TractGraphFormer | **95.59 (2.17)** | **95.57 (2.22)** | **95.65 (2.10)** | **95.56 (2.17)** |

$p<0.0005$*** $p<0.005$** $p<0.05$ *

Table 7 Ablation study of network components for age prediction on ABCD and HCP datasets.

| Models | ABCD (months) | HCP (years) |
|---|---|---|
| Graph CNN | 5.98 (0.21)* | 2.85 (0.15)* |
| Transformer | 5.96 (0.16)** | 2.91 (0.16)* |
| TractGraphFormer w/o Att. | 5.77 (0.17) | **2.77 (0.16)** |
| TractGraphFormer | **5.76 (0.19)** | 2.78 (0.12) |

$p<0.005$** $p<0.05$*

### 3.4 Interpretation results

#### 3.4.1 Weights of Graph CNN and Transformer Streams

The weights of Graph CNN and Transformer streams ($w1$ and $w2$) during feature fusion were obtained from the trained models for both tasks and both datasets. Within each fold of the five-fold cross-validation (see details in Section 2.3.6), the two weights in the corresponding model were normalized to sum to 1. Both of the weights before (Weights) and after normalization ($\text{Weights}_{\text{norm}}$) were averaged across five folds and the results are shown in Table 8 (the numbers in parentheses indicate standard deviations across five folds). As we can see, for the two tasks and two datasets, both the Graph CNN and Transformer streams have noticeable weight values, indicating that both streams made substantial contributions to the final prediction results. For both tasks and both datasets, the Graph CNN stream has larger weights than the Transformer stream, indicating a larger contribution to the prediction results.

Table 8 Weights of Graph CNN and Transformer streams for feature fusion.

| Dataset | | ABCD | | HCP | |
|---|---|---|---|---|---|
| | | $w1$ | $w2$ | $w1$ | $w2$ |
| sex classification | Weights | 0.57 (0.05) | 0.35 (0.16) | 0.95 (0.07) | 0.12 (0.04) |
| | $\text{Weights}_{\text{norm}}$ | 0.63 (0.13) | 0.37 (0.13) | 0.89 (0.04) | 0.11 (0.04) |
| age | Weights | 0.92 (0.16) | 0.09 (0.02) | 0.42 (0.03) | 0.36 (0.09) |
| prediction | $Weights_{norm}$ | 0.90 (0.03) | 0.10 (0.03) | 0.54 (0.04) | 0.46 (0.04) |

### 3.4.2 Predictive White Matter Tracts

In this section, we enable visualization of white matter fiber tracts that were found to be highly predictive across both tasks and both large datasets under study. Attention scores of each cluster were obtained from the attention module in TractGraphFormer for all testing subjects within each fold. Therefore, following the five-fold cross-validation experiments, we obtained attention scores for all subjects under study. For the sex classification task, we only kept the attention scores of subjects whose sex category was correctly predicted for accurate interpretation of results. The predictive tracts for sex classification and age prediction are identified with our proposed interpretation method (see Section 2.3.5 for details), where we first obtain the top 50 predictive clusters with the highest importance scores and then recognize all tracts to which these predictive clusters belong. The lists of identified tracts for both datasets are provided in Table 9. Overall, for both prediction tasks, we can observe that widespread regions in the WM are predictive of the sex or age of an individual. Some of the identified predictive tracts differ across datasets, as may be expected due to the developmental changes in white matter microstructure from childhood to young adulthood (Lebel et al., 2012). For the sex prediction task, as shown in bold in Table 9 and visualized in Fig. 4, thirteen predictive tracts are consistently identified across the ABCD and HCP datasets. These include 5 thalamic projection tracts (bilateral thalamo-frontal, bilateral thalamo-temporal, and left thalamo-parietal) and 8 cortico-cortical association tracts (bilateral superior longitudinal fasciculus I, bilateral superficial frontal, left inferior longitudinal fasciculus, right middle longitudinal fascicle, right uncinate fasciculus, and right superficial parieto-temporal). For the age prediction task, eleven predictive tracts are consistently identified across both ABCD and HCP datasets, as shown in Table 9 and Fig. 5. These include 3 projection tracts (bilateral thalamo-temporal and right striato-frontal) and 8 cortico-cortical association tracts (bilateral cingulum bundle, bilateral superficial occipital-temporal, bilateral superficial temporal and bilateral uncinate fasciculus).

Table 9 Identified predictive tracts for sex classification and age prediction for each dataset.

| Dataset | ABCD | HCP |
|---|---|---|
| sex classification | CB_L, CB_R, **ILF_L**, ILF_R, MdLF_L, **MdLF_R**, **SLF-I_L**, **SLF-I_R**, SP_L, **Sup-F_L**, **Sup-F_R**, Sup-PT_L, **Sup-PT_R**, **TF_L**, **TF_R**, **TP_L**, TP_R, **TT_L**, **TT_R**, UF_L, **UF_R** | AF_L, CC1, CC2, CST_L, ICP_L, **ILF_L**, IOFF_L, Intra-CBLM-PaT_L, Intra-CBLM-PaT_R, **MdLF_R**, SF_L, SF_R, SLF-II_L, SLF-II_R, **SLF-I_L**, **SLF-I_R**, SO_L, Sup-FP_L, Sup-FP_R, **Sup-F_L, Sup-F_R**, **Sup-PT_R**, Sup-P_L, Sup-P_R, **TF_L**, **TF_R, TP_L, TT_L, TT_R, UF_R** |
| age prediction | **CB_L, CB_R**, Intra-CBLM-I-P_L, Intra-CBLM-I-P_R, Intra-CBLM-PaT_L, Intra-CBLM-PaT_R, **SF_R**, Sup-FP_L, Sup-FP_R, Sup-F_R, **Sup-OT_L, Sup-OT**_R, Sup-P_L, **Sup-T_L, Sup-T_R, TT_L, TT_R, UF_L, UF_R** | **CB_L, CB_R**, MdLF_L, **SF_R**, SLF-I_R, **Sup-OT_L, Sup-OT**_R, Sup-O_L, Sup-O_R, Sup-PO_L, Sup-PO_R, **Sup-T_L, Sup-T_R, TT_L, TT_R, UF_L, UF_R** |

AF: arcuate fasciculus; CB: cingulum bundle; CC: corpus callosum; CST: corticospinal tract; ICP: inferior cerebellar peduncle; ILF: inferior longitudinal fasciculus; IOFF: inferior occipito-frontal fasciculus; Intra-CBLM-I-P_l: intracerebellar input and Purkinje tract; Intra-CBLM-PaT: intracerebellar parallel tract; MdLF: middle longitudinal fasciculus; SF: striato-frontal; SLF: superior longitudinal fasciculus; SO: striato-occipital; SP: striato-parietal; Sup-F: superficial frontal; Sup-FP:superficial frontal-parietal; Sup-P: superficial parietal; Sup-PO: superficial parietal-occipital; Sup-PT: superficial parietal-temporal; Sup-O: superficial occipital; Sup-OT: superficial occipital-temporal; Sup-T: superficial temporal; TF: thalamo-frontal; TP: thalamo-parietal; TT: thalamo-temporal; UF: uncinate fasciculus

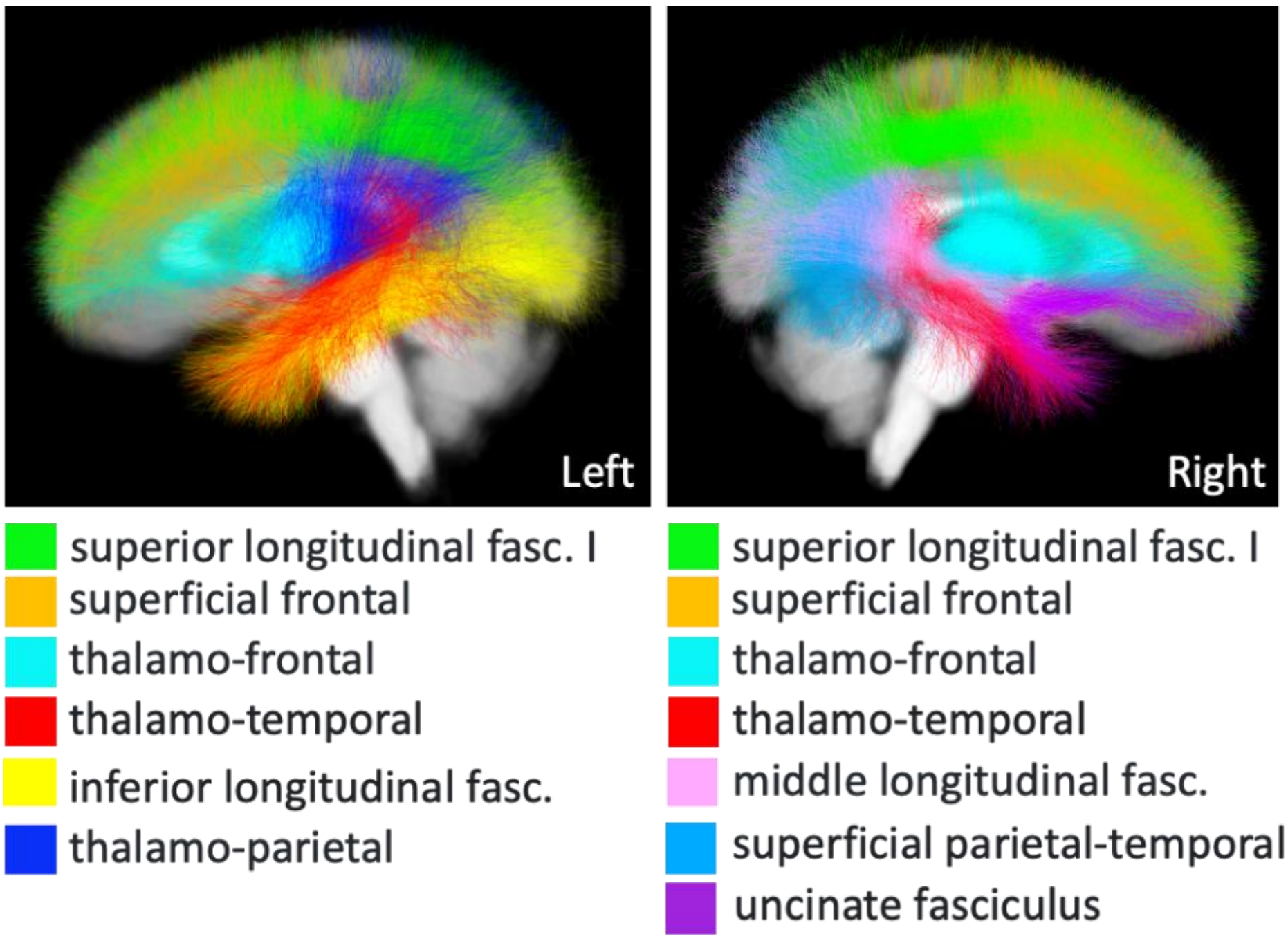

Fig. 4. Visualization of predictive tracts commonly identified across both ABCD and HCP datasets for the sex classification task.

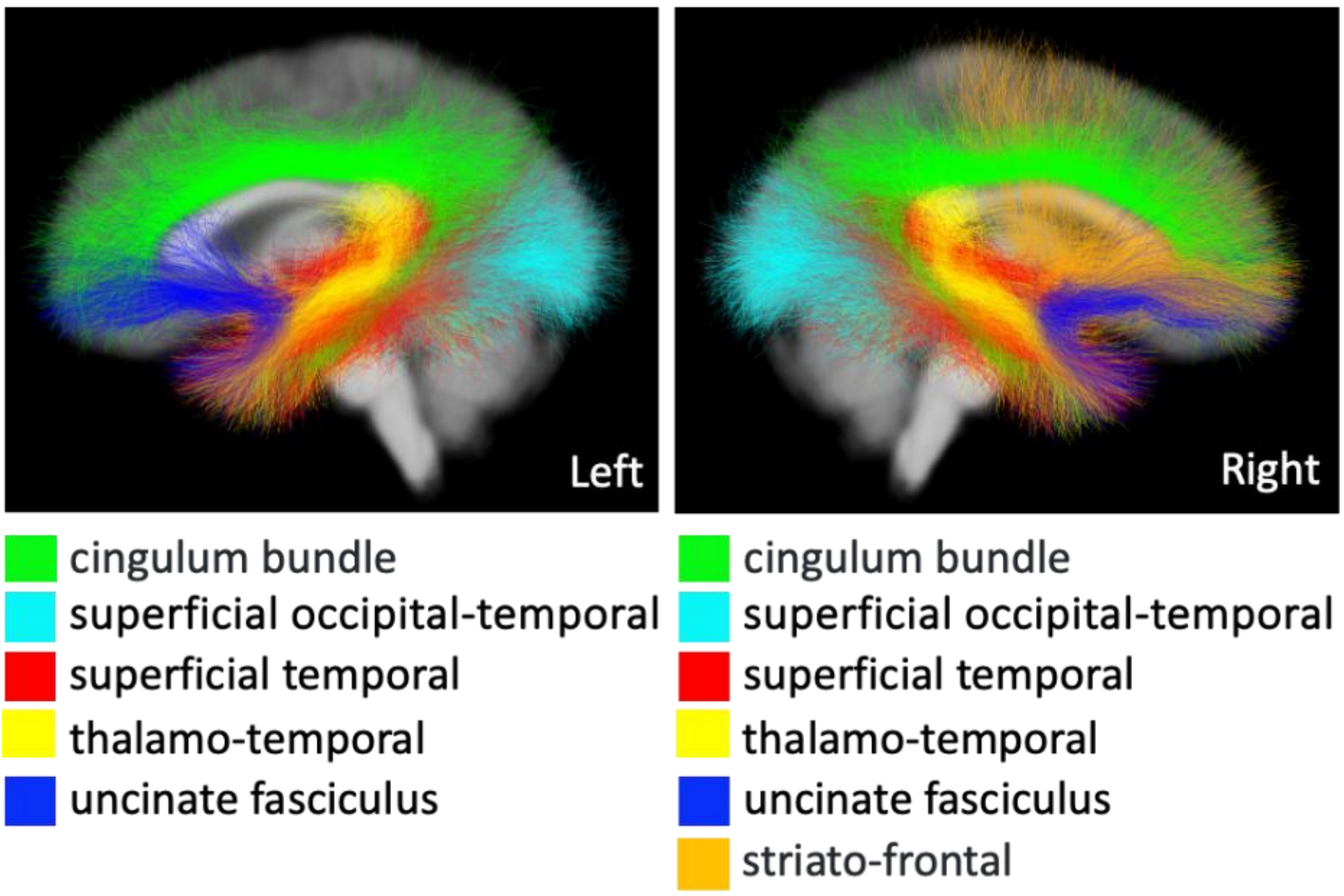

Fig. 5 Visualization of predictive tracts commonly identified across both ABCD and HCP datasets for the age prediction task.

## 4 Discussion

In this work, we propose a novel hybrid deep learning framework, TractGraphFormer, for group classification based on cluster-wise white matter features from dMRI tractography. TractGraphFormer is a two-stream deep neural network consisting of a parallel Graph CNN and Transformer feature extractor. In the Graph CNN stream, anatomically informed local features are learned by modeling the anatomical relationship between clusters as a graph. The Transformer with a self-attention mechanism extracts global features from all clusters by taking each cluster as a token. Then the local and global features are fused for classification, and an attention module enables interpretation of predictive tracts for the classification results. Our proposed method demonstrates superior performance compared to SOTA methods through evaluation on two large-scale datasets. Ablation studies show that the aggregation of anatomically informed local and global information significantly improves classification performance. A set of predictive tracts for sex classification and age prediction were successfully identified for both datasets with our method.

Our proposed TractGraphFormer framework has demonstrated superior sex and age prediction performance compared to several SOTA methods, which used the same inputs as ours. However, the studies that adopted the same inputs as ours are limited. To further demonstrate the effectiveness of our approach in achieving accurate sex classification and age prediction, we broaden our literature review to evaluate our performance in comparison to previous studies that use different white matter features such as structural connectivity. Though the direct comparison of performance is difficult due to the difference in input features, our method demonstrates competitive sex and age prediction performance compared to existing studies in the literature. For example, a recent study of sex prediction from HCP structural connectivity data achieved 92.75% accuracy (Shanthamallu et al., 2019) and the well-known BrainNetCNN achieved 76.5% accuracy in a study of 3,152 participants (Yeung et al., 2020). Another recent study of age prediction on structural connectomes from the HCP-YA dataset achieved an MAE of around 3 years (Kopetzky et al., 2024).

By comparing the sex prediction performance across the ABCD and HCP datasets, we observe that HCP obtains obviously higher classification performance than ABCD. Considering that the subjects of ABCD and HCP datasets belong to different age groups (adolescents for ABCD and young adults for HCP), the classification performance differences between the two datasets could be related to the developmental differences in white matter between males and females from adolescence to adulthood, which have been demonstrated by previous studies (De Bellis et al., 2001; Lebel et al., 2012; Schmithorst et al., 2008). As an additional assessment of model performance and generalization ability, we tested all trained models on the other, unseen dataset (i.e. we trained on ABCD and tested on HCP, and vice versa). The results indicate that this is a challenging task, likely related to the differences in dMRI acquisition and the different developmental stages of the two datasets, as mentioned above. In fact, sex-based differences in white matter are known to differ throughout the lifespan (Conte et al., 2024; Wang et al., 2012). However, we note that our proposed TractGraphFormer outperformed the compared methods in generalization across the two datasets, with the highest average accuracy of 65.73% (see Supplementary Material 1).

Our method is designed to improve performance by utilizing anatomical information that is encoded in a graph. As anatomical information, we leverage WM geometry and GM connectivity information, which have previously been shown to be useful for dMRI-based statistical analyses and machine learning prediction tasks (Yeung et al., 2023b; Zhang et al., 2018b). We build graphs where neighborhoods are defined using the anatomical information, and we adopt EdgeConv modules to aggregate features from neighboring clusters. EdgeConv (Wang et al., 2019) has previously been shown to successfully aggregate features from neighboring nodes defined in a graph (Astolfi et al., 2020; Chen et al., 2023a; Xue et al., 2023). We attribute the improved performance of our proposed graph-based networks to their ability to learn information from neighboring nodes with related anatomy. This is similar to how, in image processing, a CNN can leverage information from neighboring pixels with similar image information. In contrast, the raw cluster-wise white matter features are in no particular order, so the compared 1D-CNN method applies convolutions that combine information from anatomically different fiber clusters. The proposed graph design is also robust as it is relatively insensitive to the number of chosen neighbors in the graph. As expected, with increasing $k$ values, the prediction performance first increases due to integrating information from nearby clusters and then decreases due to including more information from distant clusters. The experimental results support our hypothesis that deep learning performance can be enhanced by incorporating information about white matter geometry and gray matter connectivity, particularly about neighborhoods of nearby and anatomically similar fiber tracts. Our results confirm the overall effectiveness of our graph model, which incorporates anatomical information to guide convolutions.

The proposed hybrid Graph CNN and Transformer framework learns local and global features simultaneously to boost prediction performance. In the computer vision community, Transformers are well known to be capable of capturing long-range dependencies, thus benefiting prediction performance (Cao et al., 2023; Lu et al., 2022; Sarasua et al., 2022; Zhang et al., 2022d). In parallel to the Graph CNN, we add a Transformer module to extract global features from all clusters, providing a supplement to the local features learned from the Graph CNN. The strategy of combining local and global information to improve performance has demonstrated success in various computer vision applications (Fang et al., 2022; A. He et al., 2023). Our experimental results show that using only the Graph CNN or Transformer stream can achieve reasonable sex or age prediction performance; however, the combination of the two streams can further improve performance significantly, due to the learned complementary local and global information. The interpretation results about the weights of the Graph CNN and Transformer streams further demonstrate that both streams contributed to the prediction results for both datasets. The Graph CNN stream plays a more important role in prediction, indicating that local features from anatomically neighboring clusters are more informative for the sex classification and age prediction tasks. The Transformer is more important for the ABCD dataset than the HCP dataset, potentially due to the fact that good Transformer performance requires large-scale training data (Yuan et al., 2021), so the Transformer is more helpful for the much larger ABCD dataset.

The adoption of an attention module enables the interpretation of predictive tracts for the sex prediction task while slightly improving prediction performance. The interpretation of these predictive tracts can allow further investigation of the relationship between sex differences and white matter connections. Several of the tracts that were found to be predictive for sex classification across the HCP and ABCD datasets have long been known to exhibit sex differences, including the inferior longitudinal fasciculus, superior longitudinal fasciculus, uncinate fasciculus, and thalamo-frontal connections (Choi et al., 2010; Jung et al., 2019; Kitamura et al., 2011). Though the effects of sex on the superficial white matter are less studied, our method identified the superficial parietal-temporal tract as a predictive tract for sex prediction. This is in accordance with one previous study that discovered sex differences in FA in the superficial white matter in temporoparietal regions (Phillips et al., 2013).

The adoption of an attention module enables the interpretation of predictive tracts for the age prediction task while maintaining similar prediction performance. In our study, five bilateral tracts (both left and right hemisphere) that connect to the temporal lobe were identified as important for age prediction across the ABCD and HCP datasets. Notably, the temporal lobe, particularly the superior temporal sulcus, is known to undergo some of the most protracted developmental changes in the human brain, using multiple different metrics (Giedd et al., 1999; Gogtay et al., 2004; Shaw et al., 2008). Late development may imply that the development of the connecting fibers is also protracted relative to other brain regions. The two studied datasets span non-overlapping age ranges (ABCD: 107-133 months; HCP: 22-36 years), capturing distinct time stages in the development of white matter fiber tract microstructure. Fractional anisotropy (FA) generally increases from infancy through adolescence, peaks in early to middle adulthood, and develops at varying rates across tracts (Lebel et al., 2012; Yap et al., 2013). The uncinate and cingulum, two tracts connecting to the temporal lobe that were found to be highly predictive in both ABCD and HCP datasets, are known to undergo large changes in FA during neurodevelopment and have longer maturation timelines, with FA peaks over age 35 (Lebel et al., 2012). So it is logical that these two tracts are predictive over the age ranges of both datasets. In addition to the predictive temporal lobe connections, one right hemisphere pathway, the striato-frontal tract, was identified. Striato-frontal connections have previously been shown to play a major role in brain aging from age 10 to 80 (Bonifazi et al., 2018).

While our main focus in this study was on network design using anatomical information, we believe this is the largest data-driven assessment of tracts that may be predictive for sex classification and age prediction, focusing on those that are in common across two large datasets. However, the interpretation of predictive white matter connections is challenging due to different scales of white matter parcellation. Due to longstanding neuroanatomical conventions, the brain's white matter fiber tracts are described and named based on their origins and terminations, as well as their course throughout the white matter (Schmahmann and Pandya, 2009). These fiber tracts vary in location, shape, and volume. The grouping of fiber clusters into named anatomical fiber tracts in our atlas was performed using an expert neuroanatomical curation process (Zhang et al., 2018c). As a result, these fiber tracts vary in volume and contain different numbers of fiber clusters. In the current project, while the original attention scores are assigned per cluster, the final visualization and interpretation results are presented at the level of anatomical tracts. Consequently, larger tracts with more clusters may appear more important during interpretation. It is unclear whether this reflects a bias towards selecting such tracts or the intrinsic significance of these tracts in the brain. The hierarchical structure of fiber clusters and tracts facilitates anatomically motivated interpretation but poses challenges. Future work may focus on developing robust methods to enhance interpretability.

The interpretation results enable the analysis of the influence of undetected clusters (where input features are set to zero) on prediction outcomes. Only a small percentage of clusters (on average, 0.24% per subject) are undetected in our dataset, due to factors including inter-individual anatomical variability. We verified that these missing clusters are not driving results. In fact, the importance scores obtained from the interpretation analysis for each prediction task have a significant inverse relationship with the number of subjects in which a cluster is missing (sex prediction: Pearson's $r$ = -0.68, p < 0.0001; age prediction: Pearson's $r$ = -0.22, p < 0.0001). Therefore, the influence of undetected clusters on prediction results is negligible.

Finally, we note some limitations of this study and directions for future research. First, we evaluated our proposed method on data with popular dMRI research acquisition protocols using the UKFt tractography method, which provided fiber-specific FA and MD microstructure measures. It is of future interest to further demonstrate the applicability of our method to other types of input data including different dMRI acquisitions, tractography algorithms, and WM features. Though fiber-specific microstructure measures such as those used here are valued for their specificity to particular anatomical bundles of the brain (Dhollander et al., 2021; A. Han et al., 2023; Zhang et al., 2018b), additional features from single-tensor or HARDI models (Chamberland et al., 2019), or white matter shape features (Lo et al., 2024a), are of interest to investigate in the future. Such an investigation is not possible in the current study due to the high computational requirements to estimate additional WM features. We note that this work is resource-intensive, as generation of the ABCD dataset employed here required about 50,000 CPU hours (Cetin-Karayumak et al., 2024). We also note that the NoS or streamline count feature, though widely used and considered the top performer for many prediction tasks (Yeung et al., 2023a), is known to vary across tractography algorithms and is influenced by parameters like the number of seed points for tractography initialization (Girard et al., 2014; Jones et al., 2013). It can also be biased by erroneous streamlines (Girard et al., 2014). In our study, this bias was mitigated by using 953 expertly curated fiber clusters identified as true positive connections, with anatomically annotated false positive fiber clusters removed prior to analyses (Lo et al., 2024a; Xue et al., 2024; Zhang et al., 2018c). Additionally, an automated outlier removal step filtered implausible streamlines from each cluster (O'Donnell et al., 2017; Zhang et al., 2018c). Second, future work may investigate the utility of incorporating other anatomical relationships, such as the proximity or connectivity of gray matter regions, to enhance network design for the analysis of additional types of brain morphometric data such as cortical volume or thickness. Additionally, other WM information, such as fiber cluster traversal of bottleneck regions (Schilling et al., 2022), could potentially be incorporated to enhance graph neighborhood definitions. Third, in the current work, we evaluated our method on two testbed tasks of sex classification and age prediction using dMRI data from two large datasets of different ages. Future work could investigate the performance of our method on data from other groups such as those related to aging or disease. Future work could also move beyond a binary prediction of sex to conceptualize sex along a continuum (Phillips et al., 2019; Sanchis-Segura et al., 2022). In addition, the statistical analysis of multiple classifiers is still an active area of research. In this work, we applied the traditional ANOVA method; however, future research could investigate alternate approaches such as confidence curves (Berrar, 2017). Finally, in the current work, we utilized a straightforward weighted sum strategy to fuse learned local and global features. Future work could explore more advanced feature fusion methods that may further improve prediction performance (Z. Han et al., 2023).

## 5 Conclusion

In conclusion, we propose an anatomically informed hybrid Graph CNN and Transformer model for machine learning with diffusion MRI tractography. This model combines Graph CNN and Transformer architectures to capture robust local and global features by leveraging anatomical relationships and long-range interactions. It demonstrates strong performance in two tasks of sex and age prediction across two large datasets. An attention module helps identify important tracts for accurate sex and age prediction, showing that widespread WM tracts are predictive and consistent across datasets. The identified tracts for sex classification are thalamic projection tracts (bilateral thalamo-frontal and thalamo-temporal, and left thalamo-parietal) and association tracts (bilateral superior longitudinal fasciculus I and superficial frontal, left inferior longitudinal fasciculus, and right middle longitudinal fascicle, uncinate fasciculus, and superficial parieto-temporal). The identified tracts for age prediction are projection tracts (bilateral thalamo-temporal and right striato-frontal) and association tracts (bilateral cingulum bundle, bilateral superficial occipital-temporal, bilateral superficial temporal and bilateral uncinate fasciculus). The TractGraphFormer approach highlights the potential of incorporating local anatomical information and global feature dependencies for improved prediction performance in machine learning using dMRI tractography.

**Declaration of Generative AI and AI-assisted technologies in the writing process**

During the preparation of this work the authors used ChatGPT in order to improve the readability of several paragraphs in the Introduction. After using this tool/service, the authors reviewed and edited the content as needed and took full responsibility for the content of the publication.

**Data and code availability**

The ABCD and HCP datasets used in this project can be downloaded through the ABCD Study (https://abcdstudy.org/) and ConnectomeDB (db.humanconnectome.org) websites. The harmonized dMRI data, extracted white matter clusters, and microstructure measures from the ABCD study are available through the NIMH Data Archive (NDA) repository (https://nda.nih.gov/edit_collection.html?id=3371) (Cetin-Karayumak et al., 2024). The code for data analysis will be available upon publication at https://github.com/AnnabelChen51/TractGraphFormer.

**Author contributions**

Yuqian Chen: Conceptualization, Methodology, Software, Writing - Original draft preparation. Fan Zhang: Conceptualization, Methodology, Writing - Review & Editing. Meng Wang: Methodology, Writing - Review & Editing. Leo R. Zekelman: Data Curation, Writing - Draft and Editing. Suheyla Cetin-Karayumak: Data Curation. Tengfei Xue: Methodology, Writing- Reviewing and Editing. Chaoyi Zhang: Methodology, Writing- Reviewing and Editing. Yang Song: Writing- Reviewing and Editing. Nikos Makris: Conceptualization, Writing- Reviewing and Editing. Yogesh Rathi: Writing- Reviewing and Editing. Weidong Cai: Resources, Supervision, Writing-Reviewing and Editing. Lauren J. O'Donnell: Conceptualization, Methodology, Writing - Review & Editing, Supervision, Funding acquisition.

## Acknowledgments

This work was supported by the National Institutes of Health (NIH) grants: R01MH132610, R01MH125860, R01MH119222, R01MH116173, P41EB015902, R01NS125307, R01NS125781, R01MH112748, R01AG042512, and K24MH116366; the National Key R&D Program of China (No. 2023YFE0118600); the National Natural Science Foundation of China (No. 62371107).